\documentclass[10pt,twocolumn,letterpaper]{article}

\usepackage{iccv}
\usepackage{times}
\usepackage{epsfig}
\usepackage{graphicx}
\usepackage{amsmath}
\usepackage{amssymb}
\usepackage[accsupp]{axessibility}

\usepackage{booktabs}
\usepackage{multirow}
\usepackage[table,xcdraw]{xcolor}
\usepackage{makecell}
\usepackage{caption}
\usepackage{enumitem}

\usepackage[pagebackref=true,breaklinks=true,letterpaper=true,colorlinks,bookmarks=false]{hyperref}
\usepackage[inkscapeformat=png]{svg}

\usepackage[capitalize]{cleveref}
\usepackage{authblk}
\usepackage{algorithm}
\usepackage{algpseudocode}

\usepackage{subcaption}
\usepackage[pagebackref=true,breaklinks=true,letterpaper=true,colorlinks,bookmarks=false]{hyperref}
\usepackage[inkscapeformat=png]{svg}

\newcommand{\ourblock}{AdaBD\xspace}
\newcommand{\ourmodel}{Shortcut-V2V\xspace}
\def\thefootnote{*}

\iccvfinalcopy 

\def\iccvPaperID{3765} 

\ificcvfinal\fi

\begin{document}

\title{Shortcut-V2V: Compression Framework for Video-to-Video Translation \\ based on Temporal Redundancy Reduction}
\author[1]{Chaeyeon Chung\thefootnote{}}
\author[1,2]{Yeojeong Park\thefootnote{}}
\author[1]{Seunghwan Choi}
\author[1]{Munkhsoyol Ganbat}
\author[1]{Jaegul Choo}
\affil[1]{KAIST AI}
\affil[2]{KT Research \& Development Center, KT Corporation}
\affil[ ]{\tt\small{\{cy\_chung, indigopyj, shadow2496, soyol, jchoo\}@kaist.ac.kr}}



\maketitle\footnotetext{indicates equal contributions.}

\ificcvfinal\thispagestyle{empty}\fi

\begin{abstract}
Video-to-video translation aims to generate video frames of a target domain from an input video.
Despite its usefulness, the existing networks require enormous computations, necessitating their model compression for wide use.
While there exist compression methods that improve computational efficiency in various image/video tasks, a generally-applicable compression method for video-to-video translation has not been studied much.
In response, we present Shortcut-V2V, a general-purpose compression framework for video-to-video translation.
\ourmodel avoids full inference for every neighboring video frame by approximating the intermediate features of a current frame from those of the previous frame.
Moreover, in our framework, a newly-proposed block called AdaBD adaptively blends and deforms features of neighboring frames, which makes more accurate predictions of the intermediate features possible.
We conduct quantitative and qualitative evaluations using well-known video-to-video translation models on various tasks to demonstrate the general applicability of our framework.
The results show that \ourmodel achieves comparable performance compared to the original video-to-video translation model while saving 3.2-5.7$\times$ computational cost and 7.8-44$\times$ memory at test time. Our code and videos are available at \href{https://shortcut-v2v.github.io/}{https://shortcut-v2v.github.io/}.
\end{abstract}

\begin{figure*}[t!]
  \centering
  
  \includegraphics[width=\linewidth]{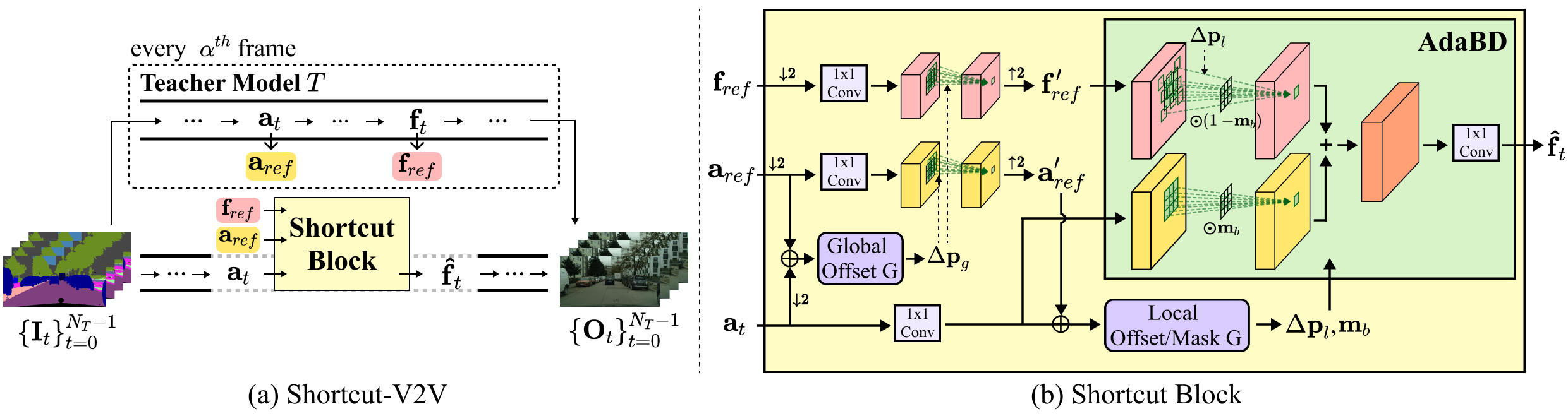}
  \caption{Overview of Shortcut-V2V. (a) is an overall framework of Shorcut-V2V, and (b) shows a detailed architecture of Shortcut block. $\uparrow2$ and $\downarrow2$ refer to upsampling and downsampling by a factor of 2, respectively. G in Offset G and Offset/Mask G indicates a generator.}
  \label{fig:architecture}
\end{figure*}

\section{Introduction}
\label{sec:intro}
Video-to-video translation is a task of generating temporally consistent and realistic video frames of a target domain from a given input video.
Recent studies on video-to-video translation present promising performance in various domains such as inter-modality translation between labels and videos~\cite{vid2vid, vid2vidfew, mallya2020world}, and intra-modality translation between driving scene videos~\cite{unsup_recycle} or face videos~\cite{recycle}.

Despite enhanced usefulness, video-to-video translation networks usually require substantial computational cost and memory usage, which limits their applicability.
For instance, multiply–accumulates (MACs) of a widely-used video translation model, vid2vid~\cite{vid2vid}, is 2066.69G, while the basic convolutional neural networks, ResNet v2 50~\cite{resnet} and Inception v3~\cite{inception}, are 4.12G and 6G, respectively. 
Furthermore, temporally redundant computations for adjacent video frames also harm the cost efficiency of a video-to-video translation network.
Performing full inference for every neighboring video frame that contains common visual features inevitably entails redundant operations~\cite{va-red, adafuse}.

In this regard, Fast-Vid2Vid~\cite{fast} proposes a compression framework for vid2vid~\cite{vid2vid} based on spatial input compression and temporal redundancy reduction.
However, it cannot be applied to other video-to-video translation models since it is designed specifically for vid2vid.
Moreover, Fast-Vid2Vid does not support real-time inference since it requires future frames to infer a current one.
Alternatively, one can apply model compression approaches for image-to-image translation~\cite{gan_compression, omgd, cat} directly to video-to-video translation, considering video frames as separate images. 
However, these approaches are not designed to consider the correlation among adjacent video frames during the compression.
This may result in unrealistic output quality in video-to-video translation, where the inherent temporal coherence of an input video needs to be preserved in the outputs.
Also, frame-by-frame inference without temporal redundancy reduction involves unnecessary computations, resulting in computational inefficiency.

In this paper, we propose \textit{Shortcut-V2V}, a general-purpose framework for improving the computational efficiency of video-to-video translation based on temporal redundancy reduction.
\ourmodel allows the original video-to-video translation model to avoid temporally redundant computations by approximating the decoding layer features of the current frame with largely reduced computations.
To enable lightweight estimation, our framework leverages features from the previous frame (\ie, reference features), which have high visual similarity with the current frame.
We also exploit current frame features from the encoding layer to handle newly-appeared regions in the current frame.
Specifically, we first globally align the previous frame features with the current frame features, and our novel \textit{Adaptive Blending and Deformation block (AdaBD)} in \ourmodel blends features of neighboring frames while performing detailed deformation. 
\ourblock adaptively integrates the features regarding their redundancy in a lightweight manner.
In this way, our model significantly improves the test-time efficiency of the original network while preserving its original performance.
\ourmodel is easily applicable to a pretrained video-to-video translation model to save computational cost and memory usage.
Our framework is also suitable for real-time inference since we do not require future frames for the current frame inference.
To the best of our knowledge, this is the first attempt at a general-purpose model compression approach for video-to-video translation. 
We demonstrate the effectiveness of our approach using well-known video-to-video translation models, Unsupervised RecycleGAN~\cite{unsup_recycle} (Unsup) and vid2vid~\cite{vid2vid}.
\ourmodel reduces 3.2-5.7$\times$ computational cost and 7.8-44$\times$ memory usage while achieving comparable performance to the original model.
Since there is no existing general-purpose compression method, we compare our method with Fast-Vid2Vid~\cite{fast} and the compression methods for image-to-image translation.  
Our model presents superiority over the existing approaches in both quantitative and qualitative evaluations.
Our contributions are summarized as follows:
\begin{itemize}
    \item We introduce a novel, general-purpose model compression framework for video-to-video translation, Shortcut-V2V, that enables the original network to avoid temporally redundant computations.
    \item We present \ourblock that exploits features from neighboring frames via adaptive blending and deformation in a lightweight manner.
    \item Our framework saves up to 5.7$\times$ MACs and 44$\times$ parameters across various video-to-video translation tasks, achieving comparable performance to the original networks. 
\end{itemize}




\section{Related Work}

\subsection{Video-to-Video Translation}
Recent video-to-video translation networks are generally classified into pix2pixHD-based~\cite{wang2018pix2pixHD} and CycleGAN-based~\cite{CycleGAN} generator.
vid2vid~\cite{vid2vid} proposes a pix2pixHD-based sequential generation framework that synthesizes a current output given the previous outputs as additional guidance. 
As the following work, few-shot vid2vid~\cite{vid2vidfew} achieves few-shot generalization of vid2vid based on attention modules, and world-consistent vid2vid~\cite{mallya2020world} is proposed to improve long-term temporal consistency of vid2vid.

While pix2pixHD-based models require paired annotated videos, RecycleGAN~\cite{recycle} and MocycleGAN~\cite{mocycle} propose CycleGAN-based video translation models.
They exploit spatio-temporal consistency losses to generate realistic videos using unpaired datasets.
STC-V2V~\cite{stc-v2v} leverages optical flow for semantic/temporal consistency to improve the output quality of the existing models.
Unsupervised RecycleGAN~\cite{unsup_recycle} achieves state-of-the-art performance among CycleGAN-based frameworks with a pseudo-supervision by the synthetic flow.
Although the existing video-to-video translation networks achieve decent performance, they commonly demand a non-trivial amount of computational costs and memory usage.
Also, frame-by-frame inference necessarily causes redundant operations due to temporal redundancy among adjacent frames.

\subsection{Model Compression}
Model compression for video-related tasks has been actively proposed in various domains~\cite{adafuse, va-red, effi_seg,delta, dynamic_quant, effi_vsr, tsm, hypercon}, such as object detection, action recognition, semantic segmentation, and super-resolution. 
Several studies~\cite{adafuse, delta, va-red} exploit temporal redundancy among video frames to improve efficiency during training or inference. 
For instance, Habibian \etal~\cite{delta} distill only the residual between adjacent frames from a teacher model to a student to speed up the inference.
Also, Fast-Vid2Vid~\cite{fast} firstly proposes a compression framework for video-to-video translation based on spatial and temporal compression.
However, Fast-Vid2Vid focuses on vid2vid~\cite{vid2vid}, limiting its application to other video-to-video translation networks.
Also, temporal redundancy reduction via motion compensation in Fast-Vid2Vid requires the future frame to infer the current frame, which is not suitable for real-time inference.

Alternatively, model-agnostic compression methods for image-to-image translation can also be applied to video-to-video translation models.
The existing approaches for image synthesis mainly tackle channel pruning~\cite{gan_compression}, knowledge distillation~\cite{cat, omgd}, NAS~\cite{gan_compression}, etc. 
For instance, CAT~\cite{cat} compresses the teacher network with one-step pruning to satisfy the target computation budget, while OMGD~\cite{omgd} conducts a single-stage online distillation in which the teacher generator supports the student generator to be refined progressively. 
However, image-based compression methods cannot consider temporal coherence among neighboring video frames, which may induce unrealistic results in video translation tasks.
Also, performing full model inference for each video frame still poses computational inefficiency due to the temporal redundancy.


\subsection{Deformable Convolution}
Deformable convolution~\cite{DCN} is originally proposed to enhance the transformation capability of a convolutional layer by adding the estimated offsets to a regular convolutional kernel in vision tasks such as object detection or semantic segmentation. 
Besides its original application, recent studies~\cite{TDAN, quality, EDVR, sr_dcn, erdn, fdan} demonstrate that deformable convolution is also capable of aligning adjacent video frames.
TDAN~\cite{TDAN} utilizes deformable convolution to capture implicit motion cues between consecutive frames by dynamically predicted offsets in video super-resolution. 
In addition, EDVR~\cite{EDVR} stacks several deformable convolution blocks to estimate large and complex motions in video restoration.
In this paper, we also leverage a deformable convolution to adaptively align adjacent video frames in a lightweight manner. 

\begin{figure}[h]
\begin{algorithm}[H]
\small
  \caption{Shortcut-V2V Inference}\label{euclid}
  \begin{algorithmic}[1]
  \State \textbf{Input:} Input video $\{\mathbf{I}_{t}\}_{t=0}^{N_T-1}$ of length $N_T$, teacher model $T$, layer index of encoder $l_e$ and decoder $l_d$, Shortcut block $S$, max interval $\alpha$
  \State \textbf{Output:} Output video $\{\mathbf{O}_{t}\}_{t=0}^{N_T-1}$

  \For{$t = 0$ to $N_T-1$}
      \State $\mathbf{a}_t = T_{[:l_e]}(\mathbf{I}_t)$
       \If{$t \% \alpha = 0$}
            \State $\mathbf{f}_t = T_{[l_e+1:l_d-1]}(\mathbf{a}_t)$
            \State $\mathbf{a}_\text{ref}$, $\mathbf{f}_\text{ref}$ $\gets$  $\mathbf{a}_t$, $\mathbf{f}_t$
            \Comment{Update the reference features}
        \Else
            \State $\mathbf{f}_t = S(\mathbf{f}_\text{ref}, \mathbf{a}_\text{ref}$, $\mathbf{a}_t)$
        \EndIf
        \State $\mathbf{O}_t = T_{[l_d:]}(\mathbf{f}_t)$
  \EndFor
  \end{algorithmic}
  \label{alg}
\end{algorithm}
\end{figure}

\section{Shortcut-V2V}
In this paper, we propose Shortcut-V2V, a general compression framework to improve the test-time efficiency in video-to-video translation.
As illustrated in Fig.~\ref{fig:architecture}(a), given $\{\mathbf{I}_{t}\}_{t=0}^{N_T-1}$ as input video frames, we first use full teacher model $T$ to synthesize the output of the first frame. 
Then, for the next frames, our newly-proposed Shortcut block efficiently approximates $\mathbf{f}_{t}$, the features from the $l_d$-th decoding layer of the teacher model. 
This is achieved by leveraging the $l_e$-th encoding layer features $\mathbf{a}_{t}$ along with reference features, $\mathbf{a}_{ref}$ and $\mathbf{f}_{ref}$, from the previous frame. 
Here, $l_d$ and $l_e$ correspond to layer indices of the teacher model.
Lastly, predicted features $\mathbf{\hat{f}}_{t}$ are injected into the following layers of the teacher model to synthesize the final output $\mathbf{\hat{O}}_{t}$.
To avoid error accumulation, we conduct full teacher inference and update the reference features at every max interval $\alpha$.
We provide the detailed inference process of \ourmodel in Algorithm~\ref{alg}.


The architecture of our model is mainly inspired by Deformable Convolutional Network (DCN)~\cite{DCN, DCNv2}, which we explain in the next section.

\subsection{Deformable Convolutional Network}
DCN~\cite{DCN, DCNv2} is initially introduced to improve the transformation capability of a convolutional layer in image-based vision tasks, \eg, object detection and semantic segmentation.
In the standard convolutional layer, a $3\times 3$ kernel with dilation 1 samples points over input features using a sampling position $\mathbf{p}_k \in \{(-1, -1), (-1, 0), ..., (0,1), (1,1)\}$.
Given input feature maps $\mathbf{x}$, DCN predicts additional offsets $\Delta \mathbf{p} \in \mathbb{R}^{2N_\mathbf{p} \times H \times W}$ to augment each sampling position along x-axis and y-axis. 
Here, $N_\mathbf{p}$ is the number of sampling positions in a kernel, and $H$ and $W$ are the height and width of output feature maps, respectively.
For further manipulation of input feature amplitudes over the sampled points, DCNv2~\cite{DCNv2} introduces a modulated deformable convolution with $\mathbf{m} \in \mathbb{R}^{N_\mathbf{p} \times H \times W}$ consisting of learnable modulation scalars.
The deformed output feature maps $\mathbf{x}'$ by DCNv2 are defined as:
\begin{equation}
        \mathbf{x}' = f_{dc}(\mathbf{w}, \mathbf{x}, \Delta \mathbf{p}, \mathbf{m}),
        \label{eq:dcn_simple}
\end{equation}
where $f_{dc}$ indicates a deformable convolution.
Specifically, a single point $\mathbf{p_o}$ of $\mathbf{x}'$ is obtained as:
\begin{equation}
        \mathbf{x'}(\mathbf{p_o}) =
        \sum_{k=1}^{N_\mathbf{p}} \mathbf{w}(\mathbf{p}_k) \cdot \mathbf{x}(\mathbf{p_o}+\mathbf{p}_k+\Delta \mathbf{p}(\mathbf{p}_k)) \cdot \mathbf{m}(\mathbf{p}_k),
        \label{eq:dcn}
\end{equation}
where $\mathbf{w}(\mathbf{p}_k)$, $\mathbf{p}(\mathbf{p}_k)$, and $\mathbf{m}(\mathbf{p}_k)$ are convolutional layer weights, offsets, and modulation scalars between $0$ and $1$, respectively, for the $k$-th sampling position.


Taking advantage of the enhanced transformation capability, we also leverage deformable convolution to align features of adjacent frames only with a few convolution-like operations, instead of using heavy flow estimation networks.

\subsection{Shortcut Block}
As described in Fig.~\ref{fig:architecture}(b), Shortcut block $S$ estimates the current frame features $\mathbf{\hat{f}}_{t}$ given $\mathbf{a}_{t}$ and the reference frame features $\mathbf{f}_{ref}$ and $\mathbf{a}_{ref}$ as inputs:
\begin{equation}
\mathbf{\hat{f}}_{t} = S(\mathbf{f}_{ref}, \mathbf{a}_{ref}, \mathbf{a}_{t}).
\end{equation}

Our block effectively obtains rich information from $\mathbf{f}_{ref}$ via coarse-to-fine alignment referring to alignment between $\mathbf{a}_{ref}$ and $\mathbf{a}_{t}$.
Also, during the fine alignment, our newly-proposed \ourblock simultaneously performs adaptive blending of the aligned $\mathbf{f}_{ref}$ and the current frame feature $\mathbf{a}_{t}$. 
Here, $\mathbf{a}_{t}$ supports the synthesis of newly-appeared areas in the current frame.

\noindent\textbf{Coarse-to-Fine Alignment.}
To handle a wide range of misalignments between the frames, our model aligns $\mathbf{f}_{ref}$ with the current frame in a coarse-to-fine manner.
Our global/local alignment module consists of an offset generator to estimate offsets, and deformable convolution layers to deform features based on the predicted offsets. 
Following TDAN~\cite{TDAN}, an offset generator estimates sampling offsets given the adjacent frame features.
For global alignment, we first downsample the given inputs to enlarge the receptive fields of the corresponding convolutional layers in a lightweight manner. 
The downsampled $\mathbf{a}_{ref}$ and $\mathbf{a}_{t}$ are concatenated and fed into a global offset generator to generate global offsets $\Delta \mathbf{p}_g \in \mathbb{R}^{2 \times \frac{H}{2} \times \frac{W}{2}}$.
Since we only need to capture coarse movement, $\Delta \mathbf{p}_g$ includes a single offset for each kernel, unlike the original DCN.
Each offset is identically applied to all the sampling positions within the kernel.
Then, the deformed features are upsampled back to the original size to obtain $\mathbf{f}_{ref}'$ as follows:
\begin{equation}
    \mathbf{f}_{ref}' = (f_{dc}(\mathbf{w}_g, (\mathbf{f}_{ref})^{\downarrow2}, \Delta \mathbf{p}_g, \mathbf{1}))^{\uparrow2},
    \label{eq:global_dcn}
\end{equation}
where $\mathbf{w}_g$ denotes weights of global deformable convolution, and $(\cdot )^{\uparrow2}$ and $(\cdot )^{\downarrow2}$ refer to upsampling and downsampling by a factor of 2 through bilinear interpolation, respectively.
$\mathbf{1}$ indicates a vector filled with 1 so that no modulation is applied here.
While local alignment of the coarsely-aligned feature $\mathbf{f}_{ref}'$ follows the process of global alignment, the difference lies in that each sampling point of each kernel has a unique offset, and the alignment operation is conducted in the original resolution.
We leverage $\mathbf{a}_{ref}'$ and $\mathbf{a}_{t}$ to estimate local offsets, where $\mathbf{a}_{ref}'$ is the aligned $\mathbf{a}_{ref}$ which is downsampled, deformed, and upsampled with the same weights $\mathbf{w}_g$ used to synthesize $\mathbf{f}_{ref}'$.

We estimate the offsets for the decoding layer features $\mathbf{f}$ using the encoding layer features $\mathbf{a}$ under the assumption that $\mathbf{a}_t$ and $\mathbf{f}_t$ have the same structural information.
In video-to-video translation, input and output frames share the same underlying structure.
Thus, it is natural for the network to learn to maintain the structural information of an input frame throughout the encoding and decoding process.
More details are described in our supplementary materials.


\noindent\textbf{Adaptive Blending and Deformation.}
During the local alignment, we also take advantage of the current frame features $\mathbf{a}_{t}$ from the encoding layer to handle the regions with large motion differences and new objects.
To achieve this in a cost-efficient way, we introduce \ourblock, which simultaneously aligns $\mathbf{f}_{ref}'$ and blends it with $\mathbf{a}_{t}$ in an adaptive manner, as illustrated in Fig.~\ref{fig:architecture}(b) AdaBD. 

First, our local offset/mask generator predicts a blending mask $\mathbf{m}_b \in \mathbb{R}^{N_\mathbf{p} \times H \times W}$ in addition to the local offsets $\Delta \mathbf{p}_l \in \mathbb{R}^{2N_\mathbf{p} \times H \times W}$.
A learnable mask $\mathbf{m}_b$ is composed of modulation scalars ranging from 0 to 1, each of which indicates the blending ratio of the current features $\mathbf{a}_{t}$ to the aligned reference features.
While DCNv2~\cite{DCNv2} originally introduces the modulation scalars to control feature amplitudes of a single input, we leverage the scalars to adaptively blend the features from two adjacent frames considering their redundant areas.

We apply deformable convolution by adding local offsets $\Delta \mathbf{p}_l$ to sampling positions of the coarsely-aligned reference features $\mathbf{f}'_{ref}$, while the current frame features $\mathbf{a}_{t}$ are fed into standard convolutional operations.
Concurrently, blending mask $\mathbf{m}_{b}$ adaptively combines the two feature maps.
In detail, an output point $\mathbf{p_o}$ of $\mathbf{\hat{f}}_{t}$ is calculated as follows: 
\begin{equation}
    \begin{aligned}
        \mathbf{\hat{f}}_{t}(\mathbf{p_o}) = \sum_{k=1}^{N_\mathbf{p}} & \mathbf{w}_l(\mathbf{p}_k) \cdot \{\mathbf{a}_{t}(\mathbf{p_o}+\mathbf{p}_k) \cdot \mathbf{m}_{b}(\mathbf{p}_k) +\\
        &\mathbf{f}_{ref}'(\mathbf{p_o}+\mathbf{p}_k+\Delta \mathbf{p}_l(\mathbf{p}_k)) \cdot (1-\mathbf{m}_{b}(\mathbf{p}_k))\},
    \end{aligned}
    \label{eq:adadb_long}
\end{equation}
where $\mathbf{w}_l$ indicates weights of local deformable convolution.

Intuitively, the higher values of $\mathbf{m}_b$ indicate the regions where current frame features are more required.
In other words, Eq.~\ref{eq:adadb_long} can be rewritten as a summation of standard convolution and deformable convolution:
\begin{equation}
    \begin{aligned}
        \mathbf{\hat{f}}_{t} = & f_{dc}(\mathbf{w}_l, \mathbf{a}_{t}, \mathbf{0}, \mathbf{m}_b) + f_{dc}(\mathbf{w}_l, \mathbf{f}_{ref}', \Delta \mathbf{p}_l, 1-\mathbf{m}_b),
    \end{aligned}
    \label{eq:adadb}
\end{equation}
In this equation, the convolutional weights $\mathbf{w}_l$ are shared between $\mathbf{a}_{t}$ and $\mathbf{f}_{ref}'$. 
$f_{dc}$ with $\mathbf{0}$ indicates DCN with zero offsets, illustrating standard convolutional operation.
To save computational costs, we decrease the channel dimension of all input features and reconstruct the original channel size before injecting the output features into the remaining layers of the teacher network. 

\subsection{Training Objectives} 
\label{losses}
To train Shortcut-V2V, we mainly leverage alignment loss, distillation loss, and GAN losses widely used in image/video translation networks~\cite{wang2018pix2pixHD, vid2vid,CycleGAN}.
First, we adopt alignment loss $L_{align}$ to train the deformation layers in Shortcut-V2V. 
Since \ourmodel aims to align the reference frame features $\mathbf{f}_{ref}$ with the current frame, $L_{align}$ computes L1 loss between the aligned feature $\mathbf{f}^{*}_{ref}$ and the current frame features $\mathbf{f}_{t}$ extracted from the teacher model.
To obtain $\mathbf{f}^{*}_{ref}$, we align $\mathbf{f}_{ref}$ in a coarse-to-fine manner without an intervention of $\mathbf{m}_{b}$ or blending with the current features.
The alignment loss $L_{align}$ is formulated as follows:
\begin{equation}
\mathbf{f}^{*}_{ref} = f_{dc}(\mathbf{w}_l, \mathbf{f}_{ref}', \Delta \mathbf{p}_l, 1),
\end{equation}
\begin{equation}
L_{align} = \left\| \mathbf{f}_{t} - \mathbf{f}^{*}_{ref} \right\|_1. 
\end{equation}

Additionally, we employ knowledge distillation losses at the feature and output levels. A feature-level distillation loss $L_{feat}$ is applied between the estimated feature $\mathbf{\hat{f}}_{t}$ and the ground truth feature $\mathbf{f}_{t}$, while an output-level distillation loss $L_{out}$ compares the approximated output $\mathbf{\hat{O}}_{t}$ to the output $\mathbf{O}_{t}$ generated by the teacher network. The perceptual loss~\cite{zhang2018perceptual} $L_{perc}$ is also incorporated to distill the high-frequency information of the outputs.

Lastly, we utilize a typical GAN loss $L_{GAN}$ and a temporal GAN loss $L_{T-GAN}$, following the existing video-based frameworks~\cite{vid2vid, 3dtgan}. 
Temporal GAN loss $L_{T-GAN}$ encourages both temporal consistency and realisticity of the output frames. 
For the GAN losses, we consider the outputs of the teacher network as real images.

The overall objective function $L_{total}$ is as follows:
\begin{equation}
    \begin{aligned}
    L_{total} = & \lambda_{align}L_{align} + \lambda_{feat}L_{feat}+ \lambda_{out}L_{out} \\ 
    & + \lambda_{perc}L_{perc} + \lambda_{GAN}L_{GAN} + \lambda_{T-GAN}L_{T-GAN},
    \end{aligned}
\end{equation} 
where $\lambda_{align}$, $\lambda_{feat}$, $\lambda_{out}$, $\lambda_{perc}$, $\lambda_{GAN}$, and $\lambda_{T-GAN}$ are hyperparameters to control relative significance among the losses.
More details on training objectives are described in the supplementary materials.

\renewcommand{\arraystretch}{1.1}
\begin{table}[t!]
\centering
\large
\resizebox{\columnwidth}{!}{%
\begin{tabular}{lllccc}
\Xhline{2\arrayrulewidth}  
{\color[HTML]{333333} Model} & {\color[HTML]{333333} Dataset} & {\color[HTML]{333333} Method} & {\color[HTML]{333333} FVD} & {\color[HTML]{333333} MACs (G)} & {\color[HTML]{333333} Param. (M)}  \\ \hline
 &  & Original  & 1.253 & 84.61 & 7.84 \\ \cline{3-6} 
 &  & CAT  & 1.310 & 22.83 & 1.48 \\
 &  & OMGD  & 1.199  & 25.85 & 1.21 \\
 & \multirow{-4}{*}{V2C} & Ours  & \textbf{1.180} & \textbf{18.19} & \textbf{0.32} \\ \cline{2-6}
 &  & Original & 1.663 & 42.36 & 7.84 \\ \cline{3-6} 
 &  & CAT  & 1.957 & 12.02 & 1.71 \\
 &  & OMGD & 2.467 & 12.93 & 1.21 \\
\multirow{-8}{*}{Unsup} & \multirow{-4}{*}{L2V} & Ours  & \textbf{1.754} & \textbf{9.09} & \textbf{0.32}  \\ \hline
 &  & Original  & 0.186 & 2066.69 & 365 \\ \cline{3-6} 
 &  & CAT  & 0.288 & 380.55 & 24.67 \\
 &  & OMGD  & 0.264 & 485.84 & 35.81 \\
 &  & Fast-Vid2Vid  & 0.223 & 767.63 & 63.17 \\
 & \multirow{-5}{*}{E2F} & Ours & \textbf{0.209} & \textbf{359.99} & \textbf{8.29} \\ \cline{2-6} 
 &  & Original  & 0.146 & 1254.17 & 411.34 \\ \cline{3-6} 
 &  & CAT  & 0.304 & 399.36 & 71.11 \\
 &  & OMGD & 0.346 & 391.59 & \textbf{47.83} \\
 &  & Fast-Vid2Vid & 0.310 & 455.29 & 71.8 \\
\multirow{-10}{*}{vid2vid} & \multirow{-5}{*}{L2C} & Ours & \textbf{0.165} & \textbf{389.16} & 52.58 \\ \Xhline{2\arrayrulewidth} 
\end{tabular}%
}
\caption{Quantitative comparison with baselines.}
\label{tab:quant_comp}
\end{table}


\begin{table}[t!]
\tiny
\centering
\resizebox{0.9\columnwidth}{!}{%
\begin{tabular}{cllll}
\hline
\multicolumn{1}{l}{Dataset} & Method & mIoU & AC & MP \\ \hline
\multirow{4}{*}{V2L} & Original & 12.2 & 16.0 & 62.8 \\ \cline{2-5}  
 & CAT & 5.99 & 8.82 & 44.2 \\
 & OMGD & 9.76 & 13.0 & 58.7 \\
 & Ours & \textbf{11.5} & \textbf{15.2} & \textbf{61.4} \\ \hline
\multirow{4}{*}{L2V} & Original & 10.0 & 15.6 & 47.3 \\ \cline{2-5} 
 & CAT & 4.10 & 7.81 & 27.6 \\
 & OMGD & 3.42 & 6.87 & 29.5 \\
 & Ours & \textbf{9.24} & \textbf{14.9} & \textbf{43.5} \\ \hline
\end{tabular}%
}
\caption{Comparison of segmentation score for V2L and FCN-score for L2V on Unsup.}
\label{tab:seg_score}
\end{table}

\begin{figure*}[t!]
  \centering
    \includegraphics[width=\linewidth]{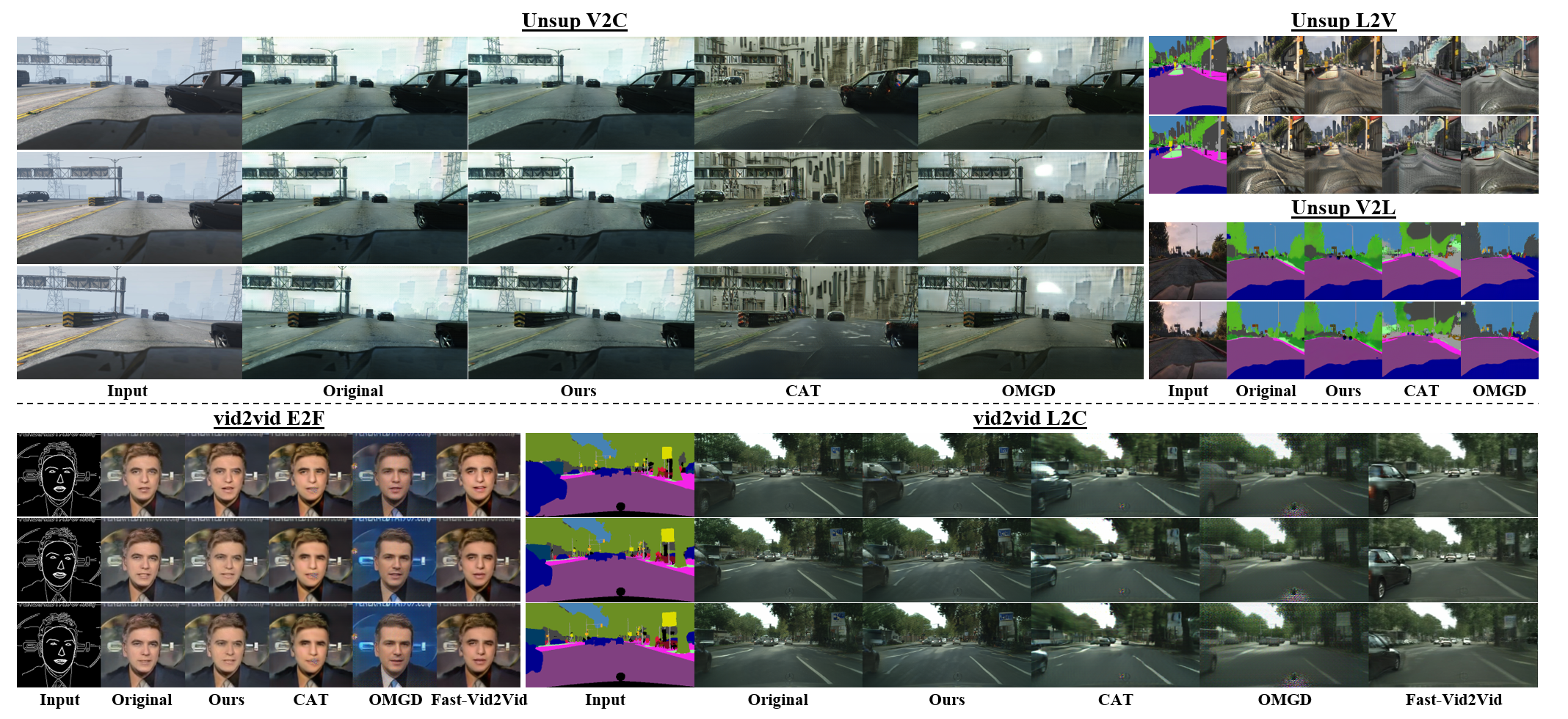}
  \caption{Qualitative comparison. The upper rows represent the results of Unsup V2C, L2V, and V2L. The bottom rows show the results of vid2vid E2F and L2C.}
  \label{fig:quali_result}
\end{figure*}

\section{Experiments}
\subsection{Experimental Settings}
\label{sec:exp_setting}
\noindent\textbf{Models and Datasets.}
To demonstrate the generality of our model, we conduct experiments on various tasks with widely-used video-to-video translation models, Unsupervised RecycleGAN~\cite{unsup_recycle} and vid2vid~\cite{vid2vid}.

Unsupervised RecycleGAN (Unsup) is the state-of-art video-to-video translation model among CycleGAN-based~\cite{CycleGAN} ones.
We conduct experiments for Unsup on Viper\textrightarrow Cityscapes (V2C), which involves translating inter-modality from game driving scenes (Viper~\cite{viper}) to real-world driving scenes (Cityscapes~\cite{Cityscapes}). 
The Viper dataset consists of driving scenes collected from the GTA-V game engine, with 56 training videos and 21 test videos.
Cityscapes is a dataset composed of 2,975 training videos and 500 validation videos. 
The input images are resized into 256$\times$512 for the experiments.
We also tackle the translation of the videos from the Viper dataset to their corresponding segmentation label maps, Viper\textrightarrow Label (V2L), and vice versa, Label\textrightarrow Viper (L2V).
We resize the images and the labels into 256$\times$256 following the previous work~\cite{recycle, unsup_recycle}.

vid2vid is a widely-used pix2pixHD-based video-to-video translation network that serves as a base architecture for various recent video-to-video translation models~\cite{vid2vidfew,mallya2020world}.
Following vid2vid, we evaluate \ourmodel on Edge\textrightarrow Face (E2F) and Label\textrightarrow Cityscapes (L2C).
E2F translates edge maps into facial videos from the FaceForensics~\cite{faces}, containing 704 videos for training and 150 videos for validation with various lengths.
The images are cropped and resized to 512$\times$512 for the experiments.
The edge maps are extracted using the estimated facial landmarks and Canny edge detector.
Also, L2C synthesizes videos of driving scenes from segmentation label maps using the Cityscapes.
We generate segmentation maps using pretrained networks following the previous studies~\cite{vid2vid, fast}.
The images and the labels are resized into 256$\times$512.

\noindent\textbf{Evaluation Metrics.}
Primarily, we adopt the Fréchet video distance (FVD) score~\cite{Unterthiner2018TowardsAG} to evaluate the performance of \ourmodel quantitatively.
The FVD score measures the Fréchet Inception distance between the distribution of video-level features extracted from generated and real videos.
The lower the FVD score is, the better visual quality and temporal consistency the generated video frames have.
For V2L and L2V, we follow the measurement of the evaluation metrics in the teacher model, Unsup~\cite{unsup_recycle}.
We measure the segmentation scores, mean intersection over union (mIoU), mean pixel accuracy (MP), and average class accuracy (AC), to validate the performance of V2L.
L2V is evaluated using FCN-score~\cite{fcn}.
For the evaluation, we first estimate label maps from the generated videos using the FCN model pretrained with the Viper dataset.
Then, we measure how accurately the estimated label maps are mapped to the ground truth segmentation labels. 
Higher FCN-scores refer to better output quality.

\noindent\textbf{Implementation Details.}
We attach our Shortcut block to the fixed teacher networks implemented based on the official codes and pretrained by the authors, except for Unsup V2C.
The teacher network of Unsup V2C is trained from scratch in the same way the original paper described~\cite{unsup_recycle}.

Also, standard convolutional kernels in the Shortcut block are replaced with HetConv~\cite{hetconv} to further enhance computational efficiency without a performance drop.
For the convenience of implementation and training stability, we intend $\mathbf{a}$ and $\mathbf{f}$ to have the same spatial size. 
Furthermore, we set the max interval $\alpha$ for each dataset considering the factors that reflect motion differences between frames, such as the frame per second (FPS) of training videos.
Lower FPS usually results in larger motion differences between the adjacent frames, requiring a shorter max interval $\alpha$ and vice versa.
Specifically, we set $\alpha$ as 3 on V2C, V2L, L2V, and L2C, since the FPS of Viper~\cite{viper} and Cityscapes~\cite{Cityscapes} is 15 and 17, respectively.
Also, $\alpha$ of E2F is set as 6, where the FPS of FaceForensics~\cite{faces} is 30.
Additional details are included in the supplementary materials.

\begin{figure}[]
  \centering
  \includegraphics[width=\linewidth]{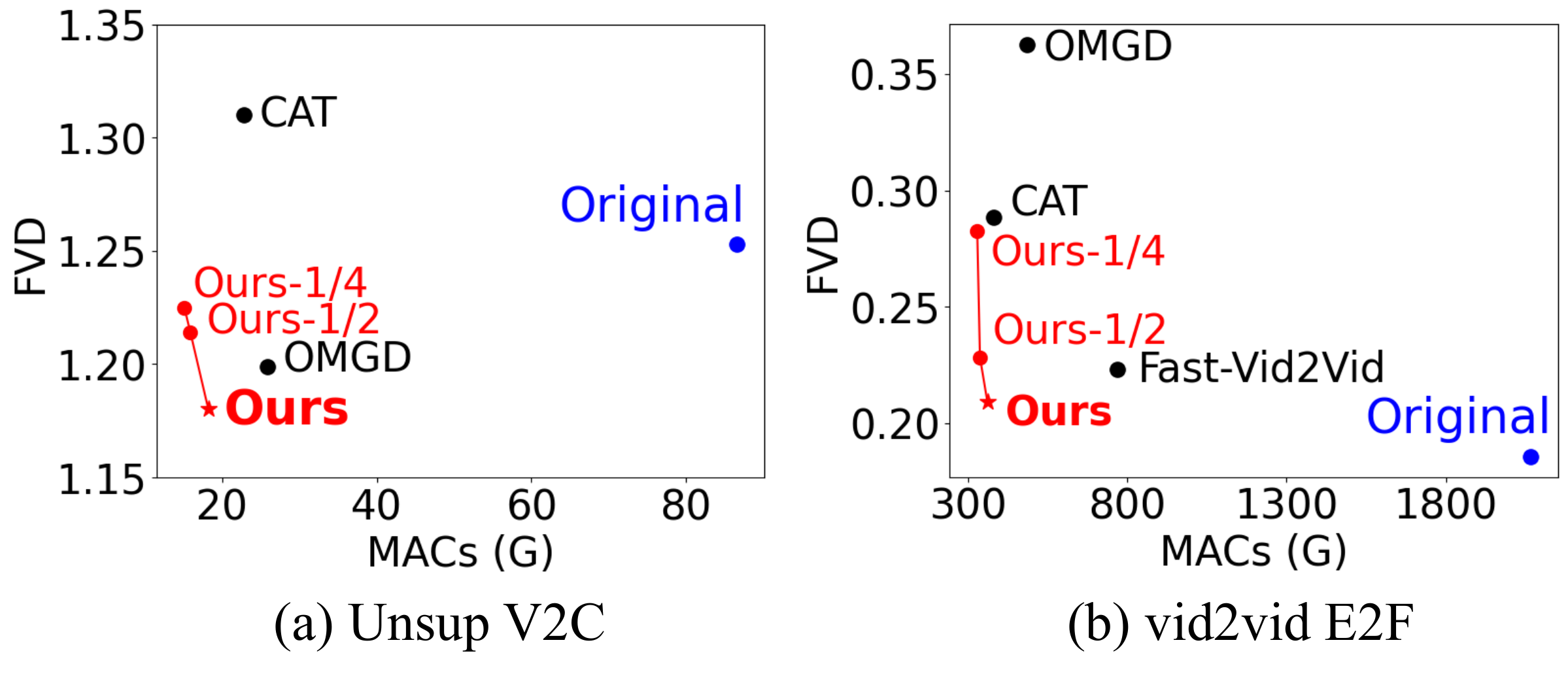}
  \caption{Performance-efficiency trade-off of the original model, Shortcut-V2V, and the existing compression methods including OMGD~\cite{omgd}, CAT~\cite{cat}, and Fast-Vid2Vid~\cite{fast}. We measure the FVD score and MACs, where the lower FVD score indicates better quality. Red points and stars denote ours with various model sizes.}
  \label{fig:tradeoff}
\end{figure} 

\subsection{Comparison to Baselines}
To demonstrate the effectiveness of our framework, we conduct qualitative and quantitative evaluations compared to the original model and other baselines.
Since this is the first work that tackles a generally applicable compression framework for video-to-video translation, we compare \ourmodel to the existing compression methods for image-to-image translation, CAT~\cite{cat} and OMGD~\cite{omgd}, regarding video frames as individual images.
In the case of vid2vid, we additionally conduct a comparison to Fast-Vid2Vid~\cite{fast}, which is the compression method designed specifically for vid2vid.
For a fair comparison, we compress the student networks of the baselines to have similar or higher MACs compared to our model.

\noindent\textbf{Qualitative Evaluation.} 
According to Fig.~\ref{fig:quali_result}, our method presents outputs of comparable visual quality to the original model with much fewer computations. 
In contrast, CAT on Unsup V2C generates undesirable buildings in the sky, and OMGD on Unsup V2C struggles with noticeable artifacts. 
For Unsup L2V, CAT and OMGD generate unrealistic textures for the terrain in the middle or the vegetation on the right.
Moreover, in V2L, CAT estimates inappropriate labels on the sky and OMGD on the vegetation.

For vid2vid E2F, CAT shows unwanted artifacts on the mouth, and OMGD presents inconsistent outputs.
For vid2vid L2C, the outputs of CAT are blurry, especially for the trees in the background, and OMGD generates artifacts at the bottom of the images.
Although Fast-Vid2Vid shows reasonable image quality, the output frames are inaccurately aligned with the input (\eg, a head pose of a person) and suffer from ghost effects (\eg, the black car on the left) due to motion compensation using interpolation.

\noindent\textbf{Quantitative Evaluation.} 
As shown in Table~\ref{tab:quant_comp} and Table~\ref{tab:seg_score}, our framework successfully improves the computational efficiency of the original network without a significant performance drop.
In the case of Unsup, \ourmodel reduces the MACs by 4.7$\times$ and the number of parameters by 24.5$\times$. 
In addition, our approach saves vid2vid's MACs by 3.2 and 5.7$\times$ and the number of parameters by 7.8 and 44$\times$ on each task.
Fig.~\ref{fig:tradeoff} visualizes the performance-efficiency trade-off of our framework and other baselines.

According to Table~\ref{tab:quant_comp}, we also outperform other compression methods for image-to-image translation, CAT~\cite{cat} and OMGD~\cite{omgd}, even with fewer computations.
Image-based approaches cannot consider temporal coherence among the frames during the compression, leading to a loss of quality.
Meanwhile, we effectively preserve the original performance by exploiting rich information in the previous features during inference. 
\ourmodel even shows superiority over Fast-Vid2Vid~\cite{fast}, which is a compression method specifically designed for vid2vid.
Table~\ref{tab:seg_score} also demonstrates that ours on Unsup surpasses the existing compression models by a large margin.

\begin{table}[]
\footnotesize
\centering
\resizebox{\columnwidth}{!}{%
\begin{tabular}{l|c|c} \hline
   Configurations & Unsup V2C  & vid2vid E2F \\  \hline
(a) w/o reference features & 1.256 &  0.398 \\
(b) w/o current features & 1.249 & 0.218 \\
(c) Single-stage alignment & 1.195 & 0.213 \\
(d) w/o adaptive blending & 1.208 & 0.244 \\ 
(e) Ours & \textbf{1.180} & \textbf{0.209} \\ 
\hline
\end{tabular}%
}
\caption{An ablation study on Unsup V2C and vid2vid E2F. We measure the FVD scores.}
\label{tab:ablation}
\end{table}

\subsection{Ablation Study} 
We conduct an ablation study to evaluate the effect of each component of Shortcut-V2V.
As described in Table~\ref{tab:ablation}, we compare the FVD scores of 5 different configurations, (a) w/o reference features, (b) w/o current features, (c) single-stage alignment, (d) w/o adaptive blending, and (e) ours, on Unsup V2C and vid2vid E2F.

First, while our model originally exploits both the reference features $\mathbf{f}_{ref}$ and the current frame features $\mathbf{a}_{t}$ to synthesize $\mathbf{\hat{f}}_{t}$, (a) and (b) are designed to leverage only either of them.
To be specific, (a) synthesizes $\mathbf{\hat{f}}_{t}$ by processing standard convolutions on $\mathbf{a}_{t}$ without integrating the deformed $\mathbf{f}_{ref}$.
Meanwhile, (b) estimates $\mathbf{\hat{f}}_{t}$ using only $\mathbf{f}_{ref}$ aligned in a coarse-to-fine manner without blending $\mathbf{a}_{t}$.
The results demonstrate that the absence of either $\mathbf{f}_{ref}$ or $\mathbf{a}_{t}$ leads to performance degradation.
Next, (c) performs a single-stage alignment on $\mathbf{f}_{ref}$ to show the necessity of a coarse-to-fine alignment.
Specifically, (c) leverages a single offset/mask generator to predict $\Delta \mathbf{p}$ and $\mathbf{m}_b$ which are then used in \ourblock for blending and deformation. 
The result indicates that coarse-to-fine alignment effectively assists the estimation of $\mathbf{\hat{f}}_{t}$ by minimizing misalignment between the features from adjacent frames.
This makes \ourmodel suitable for generating videos with diverse motion differences.
Lastly, (d) blends $\mathbf{f}^{*}_{ref}$ and $\mathbf{a}_{t}$ simply by element-wise addition instead of adaptive blending with $\mathbf{m}_b$ in AdaBD.
The result demonstrates that $\mathbf{m}_b$ also encourages better estimation of $\mathbf{\hat{f}}_{t}$ by selectively exploiting features from temporally adjacent features.



\subsection{Offset/Mask Visualization}
\begin{figure}[]
  \centering
  \includegraphics[width=\linewidth]{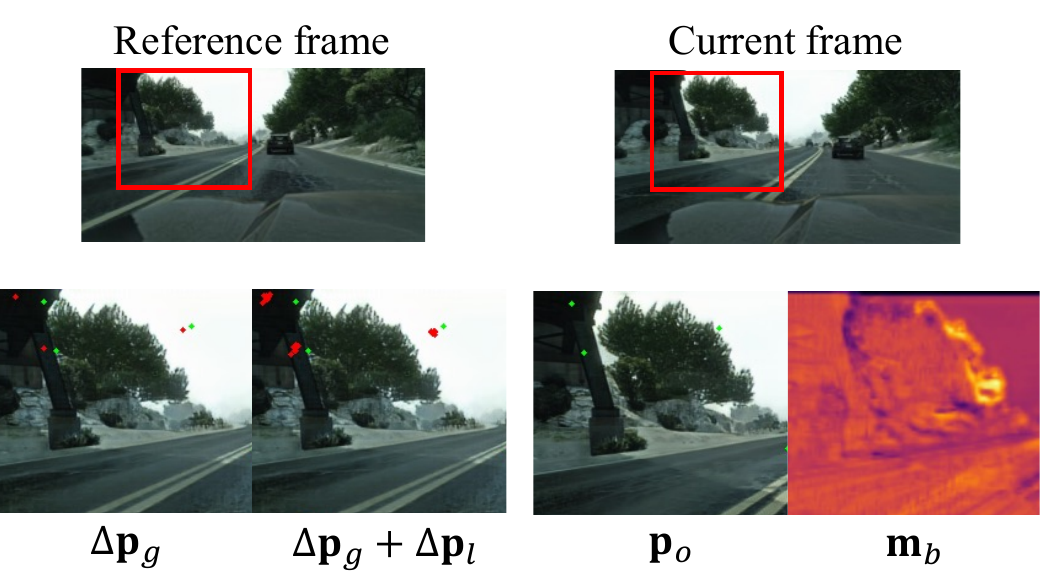}
  \caption{Visualization of global/local offsets and blending masks. The green points denote output points of the current frame, and the red points around them are each output point's sampling positions modified by global offsets $\Delta \mathbf{p}_g$ and local offsets $\Delta \mathbf{p}_l$. The values in a blending mask $\mathbf{m}_b$ are averaged by the kernel size for visualization. The brighter area indicates higher mask values.}
  \label{fig:offset_vis}
\end{figure}

We provide a qualitative analysis of the generated global/local offsets and blending masks.
Fig.~\ref{fig:offset_vis} visualizes the sampling positions (red points) for each output position $\mathbf{p}_o$ of the current frame (green points), where the predicted offsets $\Delta \mathbf{p}_g$ and $\Delta \mathbf{p}_l$ are added to the original sampling positions.
According to the result, global offsets $\Delta \mathbf{p}_g$ effectively reflect the global movement of the objects such as the bridge.
Also, the summation of the global and local offsets indicates the sampling points refined by local offsets enabling fine alignment.
This shows that the estimated offsets effectively support the utilization of common features in the reference frame.
For the blending masks $\mathbf{m}_b$, Fig.~\ref{fig:offset_vis} presents that the regions with significant motion differences (\eg, trees) have large mask values compared to regions with little change (\eg, road, sky).
That is, our model relies more on the current features rather than the reference features when the deformation is challenging, which aligns with our intention and leads to robust performance. 

\begin{table}[]
\centering
\resizebox{\columnwidth}{!}{%
\begin{tabular}{l|ccc|ccc}  \hline
\multicolumn{1}{c|}{\multirow{2}{*}{Dependence}} & \multicolumn{3}{c|}{Unsup V2C} & \multicolumn{3}{c}{vid2vid E2F} \\ \cline{2-7}
\multicolumn{1}{c|}{} & FVD & MACs (G) & Param. (M) & FVD & MACs (G) & Param. (M) \\ \hline
Low & 1.221 & 13.79 & 0.14 & 0.277 & 243.97 & 2.11 \\
\underline{Medium} & 1.180 & 18.19 & 0.32 & 0.209 & 359.99 & 8.29 \\
High & 1.166 & 27.97 & 1.38 & 0.193 & 475.99 & 32.98 \\  \hline
\end{tabular}%
}
\caption{\ourmodel performance with different teacher model dependence. High dependency denotes using more teacher network layers during the inference.}
\label{tab:depend}
\end{table}

\subsection{Performance-Efficiency Trade-off}
We present a performance-efficiency trade-off of our framework with respect to varying model sizes of Shortcut block and teacher model dependence.

\noindent\textbf{Shortcut Block Size.}
We construct our models of different sizes by reducing the number of the output channel of 1$\times$1 channel reduction layer by half (Ours-1/2) and a quarter (Ours-1/4).
As shown in Fig.~\ref{fig:tradeoff}, despite the performance-efficiency trade-off depending on channel dimension, Ours-1/2 and Ours-1/4 still achieve comparable FVD to the baselines with less computation costs.
On a single RTX 3090 GPU, our models with three different sizes on Unsup achieve 99.67, 109.69, and 111.80 FPS with an 11.6-21.2\% reduction in inference time compared to the original model's speed of 88.10 FPS. 
Additionally, Ours, Ours-1/2, and Ours-1/4 on vid2vid contribute to a significant improvement in speed from 5.63 FPS to 12.86, 12.93, and 13.16 FPS, resulting in a 56.2-57.2\% reduction in inference time. 
Our method is capable of a real-time inference, unlike the previous work~\cite{fast} which requires future frames for the current frame inference due to motion compensation.
More details on channel manipulation are described in the supplementary materials.


\noindent\textbf{Teacher Model Dependence.}
Since \ourmodel leverages a subset of the teacher network during inference, the computational costs and memory usage may vary depending on the amount of teacher network we use.
In this regard, we present an analysis of the computational efficiency and performance of our model with respect to various levels of teacher model dependence.
We categorize the dependence on the teacher model into three levels, low, medium, and high, where high dependence indicates using more teacher network layers. 
According to Table~\ref{tab:depend}, our model of higher dependence achieves better performance with larger MACs and the number of parameters.
The results also demonstrate that \ourmodel can leverage temporal redundancy in features extracted from various layers of the teacher network.
Further details for teacher model dependence and experiments for temporal redundancy are included in the supplementary materials.
\section{Discussion and Conclusion}
Although \ourmodel significantly improves the test-time efficiency of video-to-video translation, it still poses several limitations.
First, a constant max interval may induce unsatisfactory outputs when a degree of temporal redundancy largely varies between frames.
Following recent studies~\cite{stop_forward, va-red, adafuse}, applying an adaptive interval based on frame selection algorithms or a learnable policy network could be promising future research.
In addition, we need to manually configure various hyperparameters such as channel dimension, max interval, and teacher model dependence, which can be automated by NAS~\cite{gan_compression}.
Lastly, our method still has the potential to further improvement of computational efficiency by compressing the teacher model using other methods before applying our framework.

Despite the limitations, \ourmodel presents a significant improvement in test-time efficiency of video-to-video translation networks based on temporal redundancy reduction, while preserving the original model performance.
Extensive experiments with widely-used video-to-video translation models successfully demonstrate the general applicability of our framework.
To the best of our knowledge, this is the first work for a general model compression in the domain of video-to-video translation.
We hope our work facilitates research on video-to-video translation and extends the range of its application.

\vspace{0.2cm}
\noindent\textbf{Acknowledgments.} This work was supported by Institute of Information \& communications Technology Planning \& Evaluation (IITP) grant funded by the Korea government (MSIT) (No.2021-0-02068, Artificial Intelligence Innovation Hub) and the Ministry of Culture, Sports and Tourism and Korea Creative Content Agency (Project Number: R2021040097, Contribution Rate: 50).



\clearpage

{\small
\bibliographystyle{ieee_fullname}
\bibliography{egbib}
}

\clearpage

\appendix

\noindent \textbf{\Large{Supplementary Material}}
\vspace{0.3cm}

In supplementary material, we describe temporal redundancy experiments, analysis of features, implementation details, additional visualization of offsets and masks, and application of our method to the StyleGAN2-based model.

\section{Temporal Redundancy Experiments}
\begin{table}[h]
    \begin{subtable}[h]{0.45\textwidth}
        \centering
        \begin{tabular}{c|c|c}
        \hline
        Layer & CC & Norm. RMSE \\
        \hline 
        $5^{\text{th}}$ ResBlock & 0.80 & 3.917e-4\\
        $6^{\text{th}}$ ResBlock & 0.78 & 4.023e-4\\
        $1^{\text{st}}$ Upsample & 0.80 & 2.454e-4\\
        $2^{\text{nd}}$ Upsample & 0.78 & 1.667e-4\\
        \hline 
       \end{tabular}
       \subcaption{Unsup V2C}
       \label{tab:redun_unsup}
    \end{subtable}
    \hfill
    \vspace{0.3cm}
    \begin{subtable}[h]{0.45\textwidth}
        \centering
        \begin{tabular}{c|c|c}
        \hline
        Layer & CC & Norm. RMSE \\
        \hline 
        $8^{\text{th}}$ ResBlock & 0.87 & 2.341e-4\\
        $9^{\text{th}}$ ResBlock & 0.86 & 2.418e-4\\
        $1^{\text{st}}$ Upsample & 0.93 & 1.075e-4\\
        $2^{\text{nd}}$ Upsample & 0.93 & 2.813e-5\\
        $3^{\text{rd}}$ Upsample & 0.93 & 3.014e-5\\
        \hline
        \end{tabular}
        \subcaption{vid2vid E2F}
        \label{tab:redun_vid2vid}
     \end{subtable}
     \caption{Feature-level temporal redundancy between adjacent video frames. (a) and (b) are the results of Unsup V2C and vid2vid E2F, respectively. `Layer' indicates a layer which we extracted the features from. `CC' and `Norm. RMSE' denote correlation coefficient and normalized RMSE, respectively.}
     \label{tab:redun}
\end{table}

In Shortcut-V2V, neighboring video frames are assumed to have redundancy.
In this respect, we quantitatively demonstrate the feature-level temporal redundancy between adjacent video frames following VA-RED$^2$~\cite{va-red}.
VA-RED$^2$ conducts redundancy analysis to motivate its redundancy reduction framework for efficient video recognition.

As in VA-RED$^2$, we measure the Pearson correlation coefficient (CC) and normalized root mean squared error (RMSE) between adjacent frames.
CC ranges from $-1$ to $+1$, where a nonzero value indicates the existence of a positive or negative correlation, and the higher absolute value means a higher correlation.
That is, a value closer to $+1$ denotes that two variables tend to move in the same direction.
Unlike VA-RED$^2$, we normalize each feature before calculating RMSE to negate the scale differences between the features from different frames or different layers.

Since our framework approximates features from a decoding layer using the previous features from the same layer, we demonstrate the temporal redundancy of the decoding layers of the teacher networks.
Specifically, a generator of Unsupervised RecycleGAN~\cite{unsup_recycle} (Unsup) contains an input layer, 2 downsampling layers (Downsample), 6 residual blocks (ResBlock), 2 upsampling layers (Upsample), and an output layer.
For Unsup, the features are extracted from the end of the $5^{\text{th}}$ and $6^{\text{th}}$ ResBlocks, and convolutional layers of the $1^{\text{st}}$ and $2^{\text{nd}}$ Upsample.
Also, vid2vid~\cite{vid2vid} generator consists of an input layer, 3 Downsamples, 9 ResBlocks, 3 Upsamples, and an output layer.
Accordingly, we extract features from the end of the $8^{\text{th}}$ and $9^{\text{th}}$ ResBlocks, and convolutional layers of the $1^{\text{st}}$, $2^{\text{nd}}$, and $3^{\text{rd}}$ Upsample.
We use 100 randomly sampled 30-frame videos and extract 29 pairs of adjacent frames from each video. 
We conduct experiments on Viper\textrightarrow Cityscapes (V2C) for Unsup and Edge\textrightarrow Face (E2F) for vid2vid.

Table~\ref{tab:redun} presents that feature-level temporal redundancy exists regardless of the type of teacher network or layer.
Positive CC values close to 1 and normalized RMSEs close to 0 effectively justify the motivation for our approach based on temporal redundancy reduction.
Furthermore, as also presented in experiments of the performance-efficiency trade-off according to teacher model dependence (Sec. 4.5 in the main paper), our model can be applied to various layers of a teacher network.

\begin{figure}[t]
  \centering
  \includegraphics[width=\linewidth]{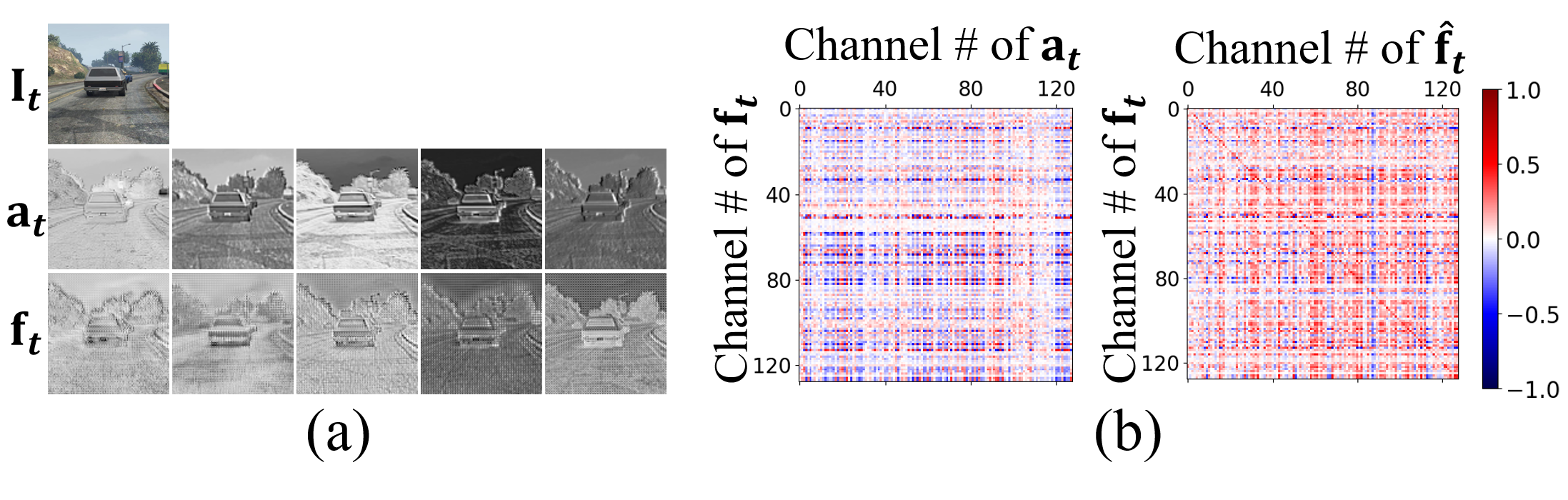}
    \caption{Visualization of (a) feature-level structural information and (b) cosine similarity map.}
    \label{fig:feat_vis}
    \end{figure}

\section{Analysis of Features}
As described in Sec.3.2 of our main paper, \ourblock predicts the offsets of feature $\mathbf{f}$ from the feature space of $\mathbf{a}$ under the assumption that $\mathbf{a}_t$ and $\mathbf{f}_t$ have the same structural information.
In video-to-video translation, input and output frames share the same underlying structure.
Thus, it is natural for the network to learn to maintain the structural information of an input frame during the encoding and decoding process.
To validate this, we randomly extract 5 channels of $\mathbf{a}_t$ and $\mathbf{f}_t$ of a random-sampled frame $\mathbf{I}_t$ from a test set. 
Fig.~\ref{fig:feat_vis}(a) shows their similar spatial structure, making it reasonable to apply offsets from $\mathbf{a}$ space to $\mathbf{f}$ space.

For blending, we note that $\mathbf{a}$ and $\mathbf{f}$ first pass through separate 1$\times$1 Convs, which serve to convert the features into the same space while reducing channel dimension.
To show the convertibility of the separate 1$\times$1 Convs, we design a network consisting of two 1$\times$1 Convs and show that it can learn to directly convert $\mathbf{a}_t$ to $\mathbf{f}$-space features $\hat{\mathbf{f}}_t$ through $L1$ loss with $\mathbf{f}_t$.
Fig.~\ref{fig:feat_vis}(b) shows the average cosine similarity maps for all channels of $\mathbf{a}_t$ and $\mathbf{f}_t$, and $\hat{\mathbf{f}}_t$ and $\mathbf{f}_t$ from 100 randomly sampled 30-frame test videos. 
The similarities between $\hat{\mathbf{f}}_t$ and $\mathbf{f}_t$ notably increase, particularly for the same channels (\ie diagonal), compared to $\mathbf{a}_t$ and $\mathbf{f}_t$.

\section{Implementation Details}
\subsection{Model Architectures}

\textbf{Coarse-to-Fine Alignment.} 
The global offset generator consists of 2 convolutional layers given two temporally neighboring features but downsampled by a factor of 2, $(\mathbf{a}_{ref})^{\downarrow 2}$ and $(\mathbf{a}_{t})^{\downarrow 2}$.
The downsampled inputs encourage the global offset generator to capture relatively coarse movements with a larger receptive field.
Meanwhile, the local offset/mask generator includes 3 convolutional layers given inputs of the original resolution, $\mathbf{a}_{t}$ and $\mathbf{a}'_{ref}$ (globally aligned features).
While local offsets contain a distinct offset for every sampling point in each kernel, global offsets include a single offset per kernel to further support the capture of coarse motions.

As mentioned in Sec. 3.2 of the main paper, we obtain $\mathbf{a}_{ref}'$ by deforming $\mathbf{a}_{ref}$, the current frame features from an encoding layer, with the same parameters used for $\mathbf{f}_{ref}$, the reference features, in Eq.~4 of our main paper, where $\mathbf{a}_{ref}' = (f_{dc}(\mathbf{w}_g, (\mathbf{a}_{ref})^{\downarrow2}, \Delta \mathbf{p}_g, \mathbf{1}))^{\uparrow2}$.
Here, $f_{dc}$ indicates a deformable convolution, $\mathbf{w}_g$ is convolutional layer weights for global alignment, $\Delta \mathbf{p}_g$ denotes global offsets, and $\mathbf{1}$ is a vector filled with the scalar value 1.

\textbf{AdaBD.}
AdaBD includes two convolutional layers, one for adaptive blending and deformation, and the other for the reconstruction of the output channel dimension.
Given a single set of convolutional weights for the adaptive blending and deformation, we apply deformable convolutions by adding local offsets $\Delta \mathbf{p}_{l}$ to the sampling position on the coarsely-aligned reference features $\mathbf{f}'_{ref}$, while the current frame features $\mathbf{a}_{t}$ is fed into standard convolutional operations without deformation.
At the same time, a blending mask $\mathbf{m}_{b}$ adaptively combines the two feature maps.
In detail, a point $\mathbf{p_o}$ of the estimated current feature $\mathbf{\hat{f}}_{t}$ obtained by Eq. 5 in our main paper. It can be expanded as follows:
\begin{equation}
    \small
    \begin{aligned}
        \mathbf{\hat{f}}_{t}(\mathbf{p_o}) = 
        &\sum_{k=1}^{N_\mathbf{p}} \mathbf{w}_l(\mathbf{p}_k) \cdot \{ \mathbf{a}_{t}(\mathbf{p_o}+\mathbf{p}_k) \cdot \mathbf{m}_b(\mathbf{p}_k) \}
         + \\
         &\sum_{k=1}^{N_\mathbf{p}} \mathbf{w}_l(\mathbf{p}_k) \cdot \{ \mathbf{f}'_{ref}(\mathbf{p_o}+\mathbf{p}_k+\Delta \mathbf{p}_l(\mathbf{p}_k)) \cdot (1 - \mathbf{m}_b(\mathbf{p}_k)) \},
    \end{aligned}
\end{equation}
where $N_\mathbf{p}$ indicates the number of sampling points in a kernel and $\mathbf{w}_l$ is the convolutional weights.
Thus, we implement AdaBD easily by a summation of convolution and deformable convolution as written in Eq. 6 in our main paper.

\begin{figure*}[t!]
  \centering
  \includegraphics[width=\linewidth]{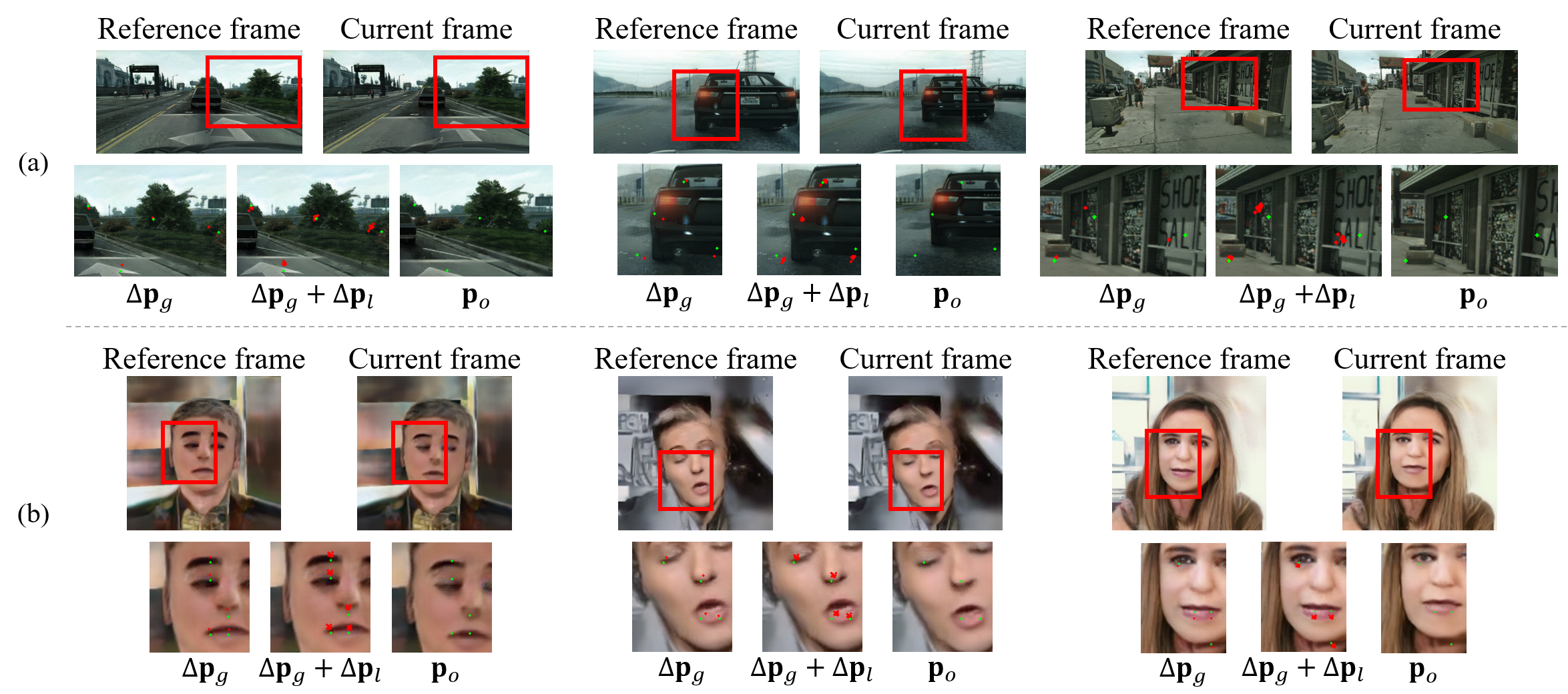}
  \caption{Additional offset visualization. (a) presents the results of Unsup V2C and (b) shows the ones of vid2vid E2F. The cropped images with $\Delta \mathbf{p}_g$ and $\Delta \mathbf{p}_g + \Delta \mathbf{p}_l$ indicate the area of a red box in reference frames. On the other hand, the cropped images with $\mathbf{p}_o$ are the region of a red box in current frames. The green points are the output points of the current frame and the red points are their sampled points.}
  \label{fig:offset_result}
\end{figure*}

\begin{figure}[]
  \centering
  \includegraphics[width=\linewidth]{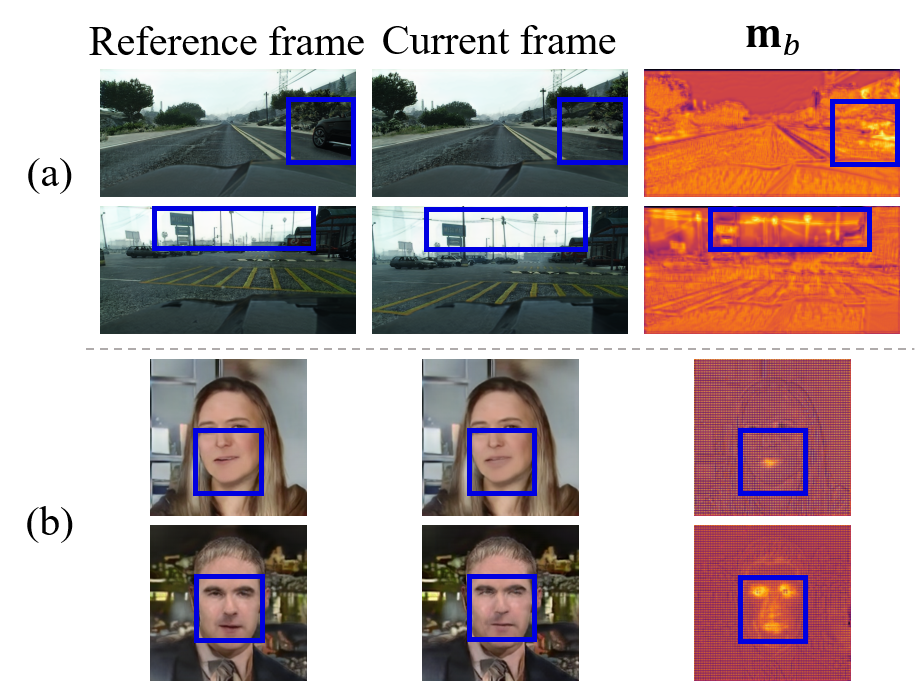}
  \caption{Additional mask visualization. The third column of (a) and (b) shows blending masks $\mathbf{m}_b$ from Unsup V2C and vid2vid E2F, respectively. The areas with blue boxes represent the regions with higher mask values compared to other regions.}
  \label{fig:mask_result}
\end{figure} 

\subsection{Details on Training Objectives}
We provide details on the training objectives we described in Sec.3.3 in our main paper.
The distillation losses, $L_{feat}$ and $L_{out}$, are computed as follows:
\begin{equation}
    L_{feat} = \left\| \mathbf{f}_{t} - \mathbf{\hat{f}}_{t} \right\|_1
\end{equation}
\begin{equation}
    L_{out} = \left\| \mathbf{O}_{t} - \mathbf{\hat{O}}_{t} \right\|_1
\end{equation}
The perceptual loss $L_{perc}$ between the teacher output and the estimated output is defined as follows:
\begin{equation}
    L_{perc} = \left\| VGG_j(\mathbf{O}_{t}) - VGG_j(\mathbf{\hat{O}}_{t}) \right\|_1,
\end{equation}
where $VGG_j$ indicates the $j$-th layer in VGG-Net\cite{vgg} and $j \in \{1,2,3,4\}$.

Finally, GAN loss $L_{GAN}$ and temporal GAN loss $L_{T-GAN}$ are formulated as follows:
\begin{equation}
    L_{GAN} = \mathbb{E}[\log D(\mathbf{O}_{t})] + \mathbb{E}[\log (1-D(\mathbf{\hat{O}}_{t})]
\end{equation}
\begin{equation}
    \begin{aligned}
    L_{T-GAN} = &\mathbb{E}[\log D_T(\mathbf{O}_{ref};\mathbf{O}_{t})] + \\
    &\mathbb{E}[\log (1-D_T(\mathbf{O}_{ref};\mathbf{\hat{O}}_{t})],
    \end{aligned}
\end{equation}
where $D$ and $D_T$ denote the discriminator and the temporal discriminator, respectively. Also, $ref$ is the reference timestep used to estimate frames at $t$ timestep.
The reference timestep's output $\mathbf{O}_{ref}$ is fed into $D_T$ in addition to the current output $\mathbf{\hat{O}}_{t}$ to encourage temporal consistency between the temporally neighboring outputs. The temporal discriminator is composed of several 3D convoluational layers based on the architecture of PatchGAN~\cite{pix2pix} discriminator.
Here, the teacher model outputs are regarded as real images.
For hyperparameters, we set $\lambda_{GAN}$, $\lambda_{T-GAN}$, $\lambda_{feat}$, $\lambda_{align}$, $\lambda_{out}$, and $\lambda_{perc}$ as 1, 1, 5, 5, 10, and 10 respectively. 

\subsection{Training and Inference}

\ourmodel can be easily applied to a fixed pretrained video-to-video translation network to improve its test-time efficiency.
For the training, we provide our model with two adjacent video frames (\ie, current frame and reference frame) whose timestep difference is smaller than the maximum interval $\alpha$.
\ourmodel with Unsup randomly samples two frames within an interval of $\alpha$ from each of the 56 training videos. 
To train our model with vid2vid, it adopts a sequential generation following the teacher model, while updating the reference frame features with the teacher model's features every $\alpha$ timestep.

The offsets are zero-initialized and the blending mask is initialized with 0.5. The learning rate of the Shortcut block with Unsup is $2 \times 10^{-4}$, and $2 \times 10^{-7}$ for discriminators.  Also, the learning rate of the Shortcut block is set as $1 \times 10^{-4}$ and $5 \times 10^{-5}$ for vid2vid E2F and vid2vid L2C, respectively.
The learning rate of discriminators is $2 \times 10^{-5}$ for both tasks.
We train our model with 3200 epochs for Unsup and 40 epochs for vid2vid.
At the test phase, we perform full model inference at every $\alpha$ to obtain the output frame and two sorts of reference features, $\mathbf{f}_{ref}$ and $\mathbf{a}_{ref}$.
For the rest timesteps, we synthesize the outputs by replacing temporally redundant operations in the original network with our lightweight Shortcut block.
In the case of Unsup, we use our model's output instead of the teacher's output at every $\alpha$ while still extracting the reference features from the teacher model.
This helps to improve the temporal consistency of videos.
In addition, following vid2vid, our model with vid2vid generates the first and second frames using a pretrained image-to-image translation model.

As in Fast-Vid2Vid~\cite{fast}, we only use the first scale generator of vid2vid.
Moreover, since our model includes alignment modules, we exclude flow estimation branch of vid2vid during the inference of our model.
Unlike vid2vid on L2C which originally utilizes an additional generator for foreground object generation, we mainly compress the generators except for the foreground model.

\subsection{Performance-Efficiency Trade-off Experiments}
\textbf{Channel Dimension.}
In Sec.~4.5 of our main paper, we construct our model with varying sizes (\ie, Ours, Ours-1/2, Ours-1/4) by manipulating the output channel dimension of $1\times1$ channel reduction layer.
In particular, for Unsup, we set the channel dimension as 128 for Ours, 64 for Ours-1/2, and 32 for Ours-1/4, where the channel dimension of the inputs is 256.
Also, while the input channel dimension of vid2vid is 512, we set the channel dimension as 256 for Ours, 128 for Ours-1/2, and 64 for Ours-1/4 for vid2vid.

\textbf{Teacher Model Dependence.}
We present the performance-efficiency trade-off with three different teacher model dependence levels, low, medium, and high, as described in Sec. 4.5 in our main paper.
A higher level indicates using more layers of the original network during the inference of our model.
Specifically, our Shortcut block replaces the operations from $2^{\text{nd}}$ Downsample in the encoder to $2^{\text{nd}}$ Upsample in the decoder, for the low level.
Medium level indicates the shortcut from $1^{\text{st}}$ Downsample in the encoder to $1^{\text{st}}$ Upsample in the decoder, which is our default mode. 
Finally, we only replace all ResBlocks with Shortcut blocks for the high level.


\section{Additional Offset/Mask Visualization}
We present further analysis of the offsets and masks estimated by our model.
Fig~\ref{fig:offset_result} visualizes the predicted global offsets, $\Delta \mathbf{p}_g$, and the sum of the global offsets and local offsets, $\Delta \mathbf{p}_g + \Delta \mathbf{p}_l$.
For visualization, we scale the offsets by the ratio of output frame resolution to feature map resolution.
The green points indicate the point $\mathbf{p}_o$ of the current frame output, and the red points are the sampled points by the offsets. 
As described in Eq.2 in our main paper, the sampled points are integrated to generate the output point of the current features.

The results show that the offsets are successfully estimated to sample proper points for the alignment of the reference frame. 
For example, the global offsets capture coarse movements of the objects such as vegetation, car, and the patterns on the wall as shown in Fig~\ref{fig:offset_result}(a). 
Additionally, Fig~\ref{fig:offset_result}(b) demonstrates that the predicted global offsets properly align the faces in the reference frame with the ones in the current frame by identifying the global movements of eyes, nose, and mouse.
Overall, the summation of global and local offsets presents that the local offsets refine the sampling positions to further capture detailed movements.

Moreover, Fig~\ref{fig:mask_result} illustrates the estimated blending masks, $\mathbf{m}_b$.
The mask values are averaged by the kernel size for visualization, where the brighter areas indicate higher mask values.
The results present that the predicted masks contain higher values for the regions of large motion differences, occlusions, and new objects, marked with blue boxes.
For these regions, our model exploits more current features than reference features to synthesize the corresponding output.












\begin{table}[t]
      \centering
      \scriptsize
    \begin{tabular}{c|cc|cccc}
    \hline
    Method & MACs &Param.  & Style          & Content        & Temporal       & Overall        \\ \hline
    Original & 515.51G & 48.46M & 0.26    &  0.24      & 0.26          & 0.27         \\ \hline
    CAT & 307.32G &  20.81M &   0.12    & 0.15          & 0.12           & 0.11          \\
    Ours & 303.26G & 15.37M & \textbf{0.23} & \textbf{0.22} & \textbf{0.23} & \textbf{0.23} \\ \hline
    \end{tabular}
    \caption{User study results of Shortcut-V2V on VToonify.}
\label{tab:vtoonify}
\end{table}

\begin{figure}[t]
  \centering
    \includegraphics[width=\linewidth]{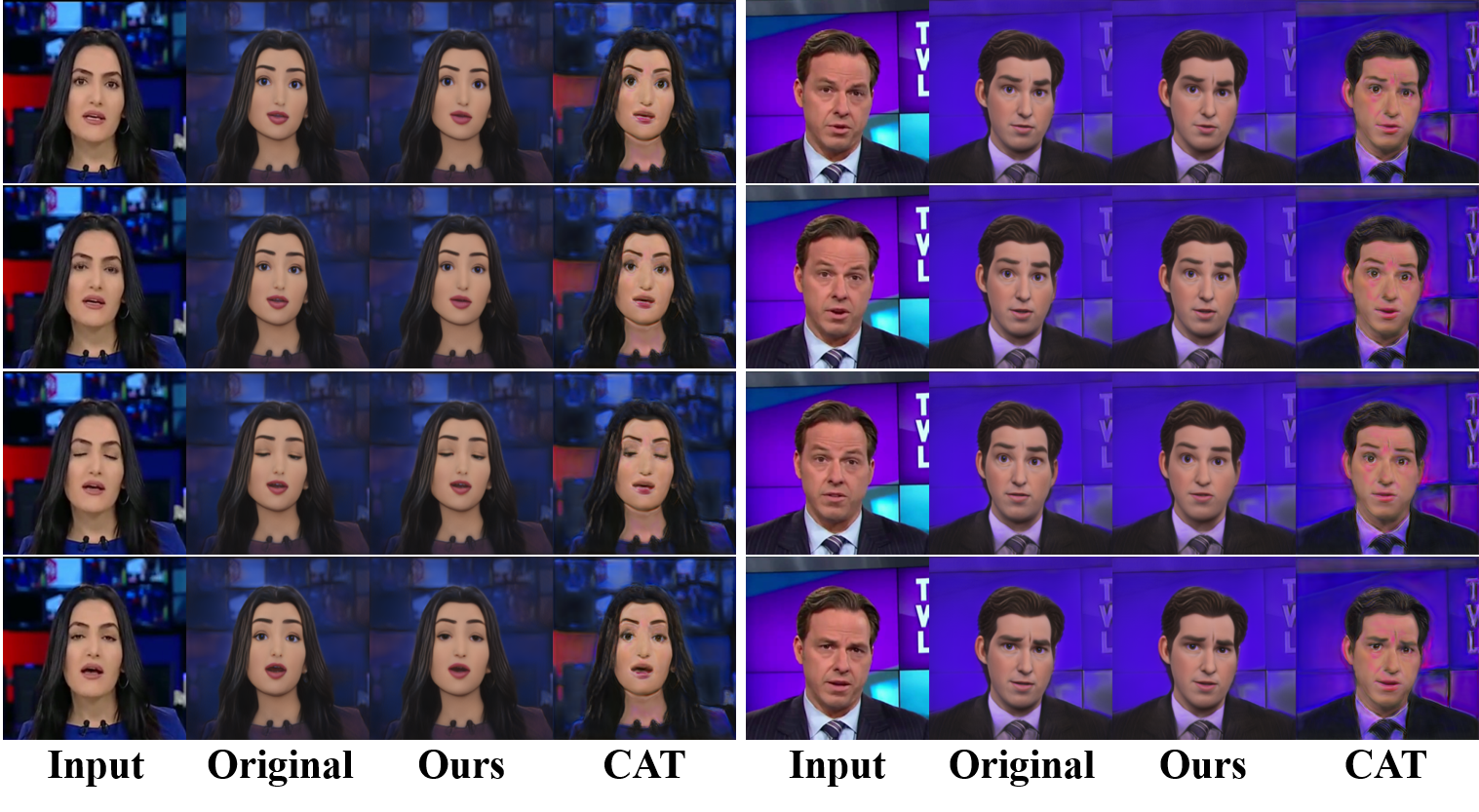}
  \caption{Qualitative comparison on VToonify.}
  \label{fig:quali_vtoonify}
\end{figure}

\section{Application to StyleGAN2-based Network}
We apply Shortcut-V2V to a recently-proposed StyleGAN2-based video-to-video translation model, VToonify~\cite{yang2022Vtoonify}.
VToonify contains an encoder and a StyleGAN2~\cite{Karras2019stylegan2} generator. Shortcut-V2V reduces computations between an encoder's final layer and the generator's initial layer. Specifically, a single layer within StyleGAN2 consists of two distinct branches: `skip' and `toRGB'. Thus, two instances of Shortcut-V2V are employed to predict each branch's features, resulting in saving 1.7$\times$ MACs and 3.2$\times$ parameters.
We reduce MACs of the original model by 41\% with Shortcut-V2V and also compare with CAT as a baseline.
For evaluation, we conduct a user study following VToonify.
Specifically, 20 graduate school students were asked to score the methods from 1 to 3 with 4 criteria: style matching, content preservation, temporal consistency, and overall quality.
In Table~\ref{tab:vtoonify}, ours obtains comparable Mean Reciprocal Ranks to the original model and significantly outperforms CAT with respect to all criteria by a large margin, demonstrating applicability to StyleGAN2-based models.
As shown in Fig.~\ref{fig:quali_vtoonify}, our method presents indistinguishable results from the original model while CAT struggles with blurry outputs.
It is worth noting that application to VToonify, a StyleGAN2-based model, provides evidence of the wide applicability of our methods across various architectures.

\clearpage

\end{document}